\documentclass[letterpaper]{article}
\usepackage{alifeconf}
\usepackage{times}
\usepackage{textcomp}

\title{Ignition: An End-to-End Supervised Model for Training Simulated Self-Driving Vehicles}

\author{Rooz Mahdavian$^{1}$, Richard Diehl Martinez $^{1}$ \\ \\
$^1$Department of Computer Science, Stanford University \\
rdm@stanford.edu}

\begin{document}
\maketitle

\begin{abstract}
We introduce Ignition: an end-to-end neural network architecture for training unconstrained self-driving vehicles in simulated environments. The model is a ResNet-18 variant, which is fed in images from the front of a simulated F1 car, and outputs optimal labels for steering, throttle, braking. Importantly, we never explicitly train the model to detect road features like the outline of a track or distance to other cars; instead, we illustrate that these latent features can be automatically encapsulated by the network. 
\end{abstract}
\section{Introduction}
\subsection{Motivation}

The current literature on self-driving vehicles has mostly focused on constrained scenarios (like street or highway driving), where the restrictions on speed and road curvature are intended to generalize to all cars. We choose, instead, to focus on the competitive motorsport, in which the unconstrained nature of driving necessitates driving policies which have considerably more variance (in optimal values for throttle, braking, and steering). Typically, variants of reinforcement learning have been applied to learn optimal self-driving policies, both in constrained and unconstrained scenarios. 
In this paper, we decide to reformulate this as a purely supervised task. In particular, we illustrate how a complex action space (the intersection of  steering, throttle, and braking) can be learned even in a supervised setting, and importantly, purely from noisy, real-world sensor data (images captured from the hood of the vehicle, and the rotation rate at the wheels of the vehicle).

Due to obvious resource constraints, we are not be able to develop a solution in a real-world setting. Fortunately, a number of extremely sophisticated simulation engines exist. 

\subsection{Simulator}
We choose to target Assetto Corsa, which is well-known for the accuracy and realism of its underlying physics engine. But more importantly, Assetto Corsa provides a sophisticated AI (which attempts to follow pre-defined splines for a given track while responding to player actions, through access to the internal simulator state). And crucially, the AI is restricted to the same physics model that the player uses (whereas most commerical simulators simplify the physics model for the AI, which in-turn dramatically simplifies it's development) \cite{AssettoCorsa}. Further, Assetto exposes internal telemetry data via a Python API, allowing to developers to write dataloggers within the game environment, which then allows us to synthesize data using the in-game AI.

\begin{figure}[h!]
    \centering
    \includegraphics[scale=0.2]{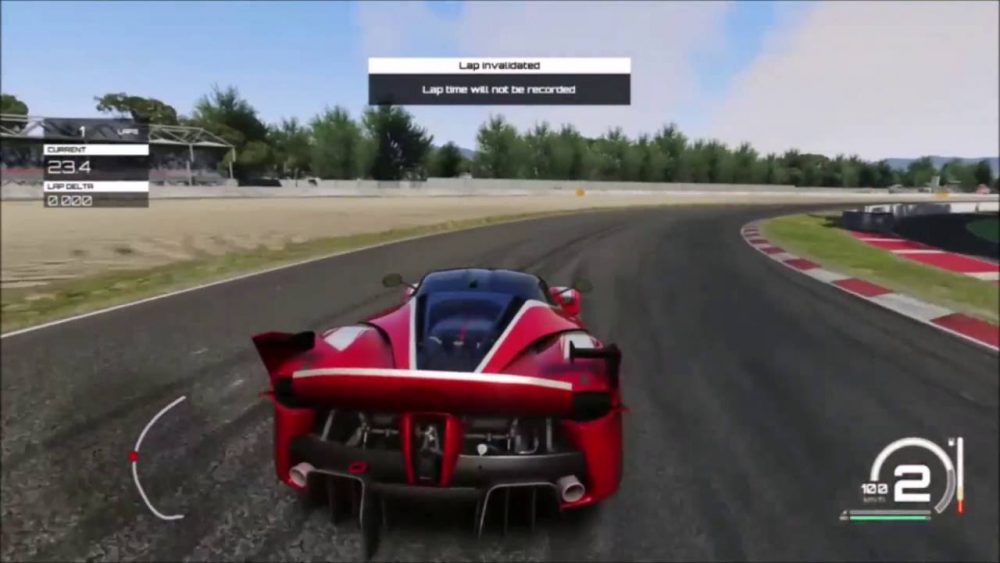}
    \caption{Assetto Corsa Gameplay}
    \label{fig:my_label}
\end{figure}

\section{Related Literature}

The idea of formulating the task of learning self-driving policies in a supervised fashion is by no means a new one. Robot mimicry, in particular, has long attempted to teach robots to navigate around spaces and perform actions by mimicking a set of instructions \cite{Hayes1994} \cite{Argall2009}. Historically, in fact, imitation learning was long employed as a standard framework for the task of learning self-driving policies. A 2004 paper by the Defense Advanced Research Projects Agency (DARPA), for instance, sought to train an end-to-end model that took in images from stereo cameras and produced throttle position and steering wheel angle as output \cite{LeCun:04}. For this task, data collection was performed by human drivers who drove a test vehicle remotely. The data from the video cameras and the steering wheel angle where then used within a supervised learning task to predict general optimal steering angles. 
In most recent years, the majority of work within the realm of learning self-driving policies has relied on reinforcement learning. Particularly with the advent of deep learning, institutions like Google and Stanford have invested heavily in deep reinforcement learning to move towards fully (class 5) autonomous vehicles \cite{Levinson2011} \cite{SebastianThrun2010}. 
Despite these advancement in reinforcement learning there exists a resurgent interest to combine more classical supervised imitation learning with deep reinforcement learning. The use of both imitation learning and reinforcement learning was first proposed by Taylor et al. in 2011. The group suggested that policy learning could be initialized with a preceding imitation learning stage \cite{Taylor2011IntegratingRL}. Doing so showed an acceleration of the reinforcement learning process and minimized costly agent-environment interactions. In addition, deploying reinforcement learning after imitation learning compensated for some of the noisy inputs that naturally arose from having a human demonstrate the desired behavior. 
In line with this research, Ross et. al developed the DAGGER algorithm in 2010 to facilitate the process of imitation learning. The group suggested training a weak model from human input and then run the controller and sample observations from the resulting trajectories of the model \cite{Ross2010}. After doing so, a human could manually label the sampled observations.
Since then, these advancements in imitation learning have been applied to self-driving algorithms. In 2016, for instance, a group of engineers at NVidia trained a deep end-to-end convolutional neural network (CNN) that learned steering commands from only a single-front facing camera \cite{Bojarski2016}. The model was trained on 72 hours of driving data, recorded on both highways and smaller roads in a variety of different weather conditions. The model illustrated that with only limited training data from humans, a CNN was able to generalize driving policies, allowing the test vehicle to operate on a number of different lanes and road types. Particularly striking was that the model was never explicitly trained to use lane-detection or path planning. Rather, all of these features were implicitly optimized simultaneously in the end-to-end nature of the model. Using this framework, the group was able to train a self-driving car which could be driven fully autonomously 98\% of the time. 
Within a simulated setting, Assetto Corsa has previously been used for training self-driving policies. Most notably, Hilleli et. al developed a framework that used only in-game images to effectively predict steering angles \cite{Hilleli2016}. During training, no access to the simulator internals were made available to the model. Importantly, however, the group relied on a so-called safety network, for preventing catastrophic mistakes in the reinforcement learning stage. This network effectively limited the viable action space that could be predicted by the final model. Notably, their neural network framework represented a synthesis of both imitation learning and reinforcement learning, relying on five models: unsupervised feature learning, supervised imitation learning, supervised reward induction, supervised module construction and reinforcement Learning.
In a similar fashion, a group from last years' CS 231 class developed a real-time Mario Kart autopilot which learned how to train and play Mario Kart without human intervention \cite{Ho2016}. The model used two main components: an omniscient AI with complete control of the emulator. This aspect of the model could simulate different possible actions and could generate a training set of associated screen shots with steering angles. 
The second aspect of the model was a convolutional neural networks which was trained on the results of the simulated dataset. One of the problems the group encountered in the training process was that the AI would occasionally run into walls and be unable to recover. In order to further augment the dataset and circumvent these issues, the group would sample states from the CNN during real-time playing. 
The model however was simplified by the fact that the autopilot was constrained to hold a constant maximum acceleration, since the game can be completed without ever breaking. Assetto Corsa, by comparison, introduces a number of additional problems in this domain since the game has a realistic physics engine which necessitates vehicles to break in order to effectively finish the race. 
At the same time, however, we realize that a number of the difficulties raised by authors in our literature review are not applicable in our particular setting. This is because the labels attached to each of the training images are retrieved directly from the in-game oracle, with does not suffer from inherent noisy behaviors. That is to say, if we frame our problem of one where we attempt to imitate the behavior of the in-game deterministic self-driving policy, then there should be no hypothetical limit to how well our model can approximate a perfect self-driving policy. We are thus optimistic that we can achieve good results by formulating our problem as a purely supervised one. Within the framework of supervised learning, we decided to implement a ResNet18 for factors that are further enumerated in section (4) relating to initial problems of over-fitting \cite{He2015}.

\begin{figure*}
    \centering
    \includegraphics[scale=0.4]{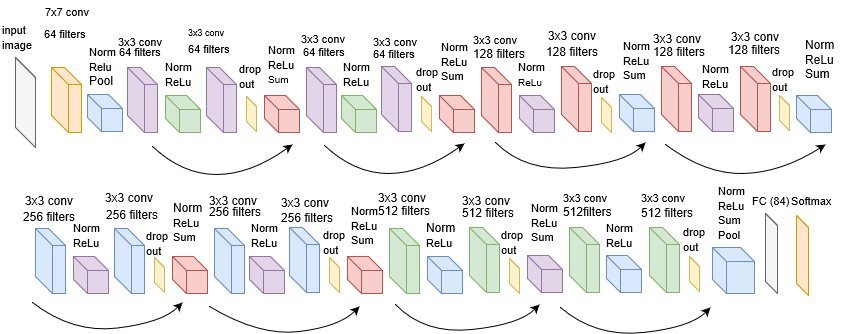}
    \caption{ResNet18 Model}
    \label{fig:my_label}
\end{figure*}

For future work, we have also considered implementing some of the suggestions listed by Chi and Mu for training self-driving policies. The authors argue that by incorporating LSTM and Conv-LSTMs into a CNN network that much better accuracy can be achieved in predicting optimal steering policies \cite{Chi2017}. Doing so, they illustrate that self-driving systems can be made to be stateful, that is to say these algorithms can incorporate information from past experience. We agree that this suggestion would allow our model to more closely approximate how real driving decision are made by humans.

\section{Data Collection}

We engineered a data synthesizer around Assetto Corsa's oracle AI, which consistent of three components: (1) a Python controller, which automates the process of selecting a pre-defined race configuration, starting the race, and handing off control to the AI, which then starts (2) an Assetto Corsa Python integration, which captures real-time telemetry data from Assetto during a race at ~100Hz, downsamples this data to 10Hz, and transfers it via RPC to (3) a separate Python process, which (for a given incoming sample) uses multiprocessing queues to capture a screenshot, and write the combination of screenshot and label to disk.

This pipeline was then instrumental in our experimentation, as we were to synthesize and experiment with a number of different datasets (which are outlined in Section (4)). Our final dataset consisted of ~8 hours of racing data with 24 different cars on the track, which amounts to over 285,000 labeled examples (~11GB worth of image data, downsampled from the game's 1280x720 resolution to a 320x180 resolution, to maintain the aspect-ratio).

We split this synthesized dataset into a training, validation and test set, with a 92-4-4 split, amounting to 265,000 images in the training set and 10,000 in both the validation and test set. Before being fed into our model, each of the images is transformed into gray-scale and normalized. 

In addition to normalizing and grey-scaling our input images, we also utilize a number of methods in order to minimize the effects of noise inherent in the input images. These data augmentation and transformation are explained in the methods and experiments section.

\begin{figure}[h!]
    \centering
    \includegraphics[scale=0.5]{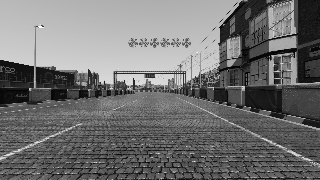}
    \caption{Braking: 0, Acceleration: 0.850, Steering: 0 }
    \label{fig:my_label}
\end{figure}
\begin{figure}[h!]
    \centering
    \includegraphics[scale=0.5]{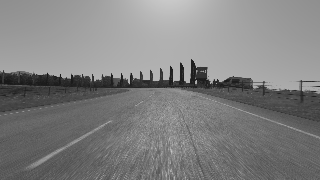}
    \caption{Braking: 0.000, Acceleration: 0.859 Steering: -13.600}
    \label{fig:my_label}
\end{figure}

\section{Methods and Experiments}
All of our experiments were trained locally, on an NVIDIA 1070 GPU, and all architecture was implemented in PyTorch.

\subsection{Baseline}
Our experimentation began with a baseline goal in mind: learn a policy for a particular track and car configuration, which mimics the oracle AI policy exactly. We synthesized a baseline dataset of \texttildelow 100,000 images (with the same track and car, against 6 opponents), at the game's native 3-channel 1280x720 resolution, and trained a regression model identical to the one described by Bojarski et. al. \cite{Bojarski2016}, but one that regresses onto values for throttle, braking, and steering (as opposed to just steering). This initial model achieved a relatively low MSE loss of 0.112 after 5 epochs (\texttildelow 6 hours), but this figure turned out to be extremely high in context: values for throttle, braking, and steering were fixed to be in the range [0,1], and thus a squared loss of 0.112 was extremely high. Moreover, the model had issues overfitting a small number of examples (\texttildelow 25), which suggested a more fundamental problem. 
\subsection{Overfit}
We hypothesized that it might have to do with numerical stability, and consequently rescaled and recentered our target values to be in the range (-100, 100). This alone allowed the model perfectly overfit 100 examples, and achieve an incredibly low loss of 0.05 (considering the new scale of the values) on 1000 examples. We then trained the model on the same baseline dataset for 5 epochs (again, \texttildelow 6 hours), but the final loss of \texttildelow 100 across the entire dataset was considerably worse. Still, this amounts to an average error of  \texttildelow 10 from our target values, and this appeared to be acceptable: 10 degree differences in steering are likely minimal mistakes, and 10 percent differences in throttle and braking input should barely make in difference at the simulator level. So, we turned our attention to building a real-time controller, to both qualitatively and quantitatively measure our results.
\subsection{Controller}
The resulting controller architecture has 3 components: (1) a Screenshotter, which dynamically finds and captures the simulator window position and size using the Windows API, and passes it to (2) Model instance, with parameters loaded from the most recent checkpoint, which performs any necessary image transforms, and computes a forward pass, handing off the resulting predictions to (3) a Joystick, which rescales the predictions to the intended in-game values, and pushes them to Assetto Corsa via pyVjoy. 
Unfortunately, the performance of the baseline model in the real-time environment was extremely poor. It would be very difficult to get it to throttle in the first place (from a start), where it consistently applied some minimal braking. Manually moving the car forward ourselves would result in it eventually making forward progress on it's own, but it would rarely be able to turn past the first corner in track without accelerating into a wall, and consequently braking and locking in place. In one instance, however, the car barely made the turn, and proceeded to drive surprisingly smoothly down a stretch of the resulting highway, which involved gradual turns that it successfully made. This was a very short stretch, however, and after \texttildelow 1000 feet it accelerated off the road, and did not recover. This was promising, but highlighted fundamental issues in the model.
\subsection{Velocity}
We then focused on the first issue highlighted above: an inability to throttle from the starting line. We hypothesized that this was due to an incorrect assumption about the nature of the problem: in extending previous work from a regression on steering to a regression on steering, throttle, and braking, we had assumed that these values were all cross-correlated given an input image state; intuitively, if an F1 car is approaching a corner, that would then imply a particular range values of values for all throttle, braking, and steering. But this picture, we believe, was incomplete: all of these values are cross-correlated conditioned on both the input image \textbf{and} the current velocity of the car. As an example, if the car is it at rest, it should always accelerate, even if it is right next to the corner; if the car is moving at a high speed, it should almost certainly apply the brakes before entering the corner; and if a car is moving at a low speed, it should likely neither accelerate nor brake, but coast. 

To this end, we synthesized a new dataset of equal size, but which now included the current velocity (in MPH) of the car as an input feature for each example. And in an effort to simplify the problem, the new dataset did not contain other cars in the road. We then modified our baseline architecture to our velocity architecture, which now concatenated the labeled velocity with the CNN-encoded state to produce a holistic state representation, which was then propagated through the same number of FC layers. We trained this new architecture on the new dataset, but we still achieved a similarly high loss (\texttildelow 100). We then evaluated the model in realtime using the controller, and found that it did, in-fact, consistently accelerate from the starting line. But the throttle and braking values were nonsensical; it would apply both throttle and brakes at the same time, but in median values (\texttildelow 0.4-0.6 for both) which resulted in stuttering motion which frequently locked up the car. We then turned to visual analysis to further refine the model.

\subsection{Visualizer}
\begin{figure}[h!]
    \centering
    \includegraphics[scale=0.2]{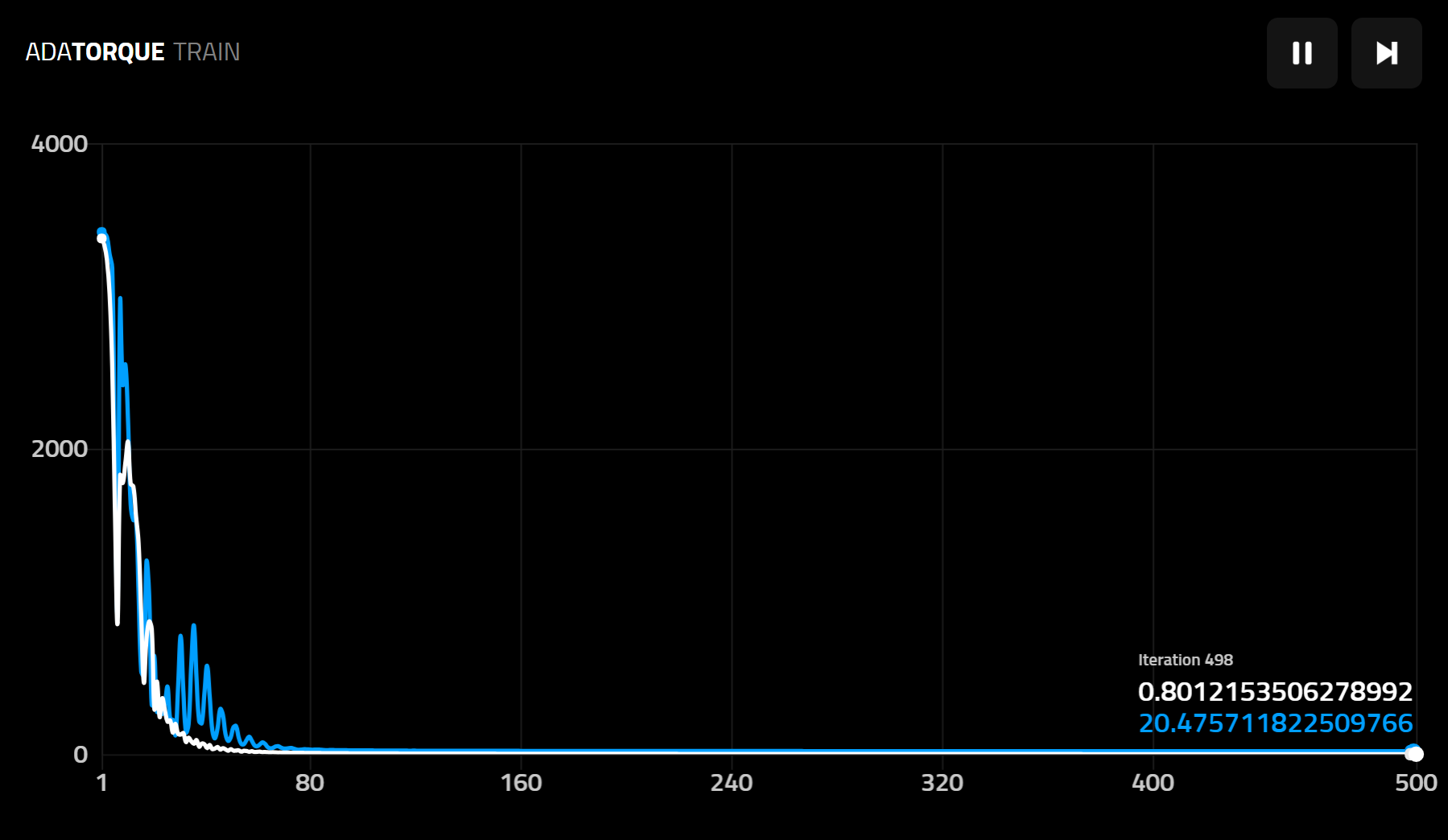}
    \caption{Example of Visualizer}
    \label{fig:my_label}
\end{figure}

To debug the model, we developed a real-time visualization architecture, which runs in a separate Python process and receives information from both the Model and the Controller via RPC, encodes the information, and relays it via JSON to a websocket server, which is then received and visualized via Javascript and HTML. The Visualizer provides both training and evaluation modes. In evaluation mode, the visualizer displays the transformed realtime input to the model (from the Assetto Corsa window), a corresponding saliency map, the model's predictions, and the corresponding being reported by the Simulator. In training mode, the Visualizer graphs realtime train and validation loss curves using Chart.js, and visualizes sampled predictions (as the input image, the predictions, and the ground truth labels). The Visualizer also supported "steps", in which realtime input could be paused (and later updates streamed in would then be cached) and each frame and it's corresponding predictions could be stepped through, one at a time.   
We then again ran trained velocity architecture in realtime, but visualized the computations in Visualizer; we chose to give the Oracle AI control, and then allowed the model make predictions on the corresponding frames, and stepped through input to see if we noticed any strange behavior. Interestingly, the values it predicted for steering were typically as close we expected (within 20 degrees) to those chosen by the Oracle AI; but the values selected for throttle and braking were substantially off.The model almost always predicted a median value for both throttle and braking, whereas the Oracle AI operated almost exclusively at the extremes of 0 or 1. Moreover, the AI (for obvious reasons) never selected a nonzero value for both. 
The corresponding saliency maps were also interesting. Initially, we computed saliency maps which visualized activations that contributed to a high ($\geq$ 0.5) throttle. And the resulting visualizations were  indistinguishable from noise, suggesting that even though we had an acceptable loss across the training set (which should be more or less identical to running it in realtime) it was entirely fitting the rendering noise and artifacts that do distinguish frames, and not learning useful information for throttle and braking.
To confirm this, we then graphed the distributions in the dataset for throttle, braking, and steering, which are provided below.
\begin{figure}[h!]
    \centering
    \includegraphics[scale=0.4]{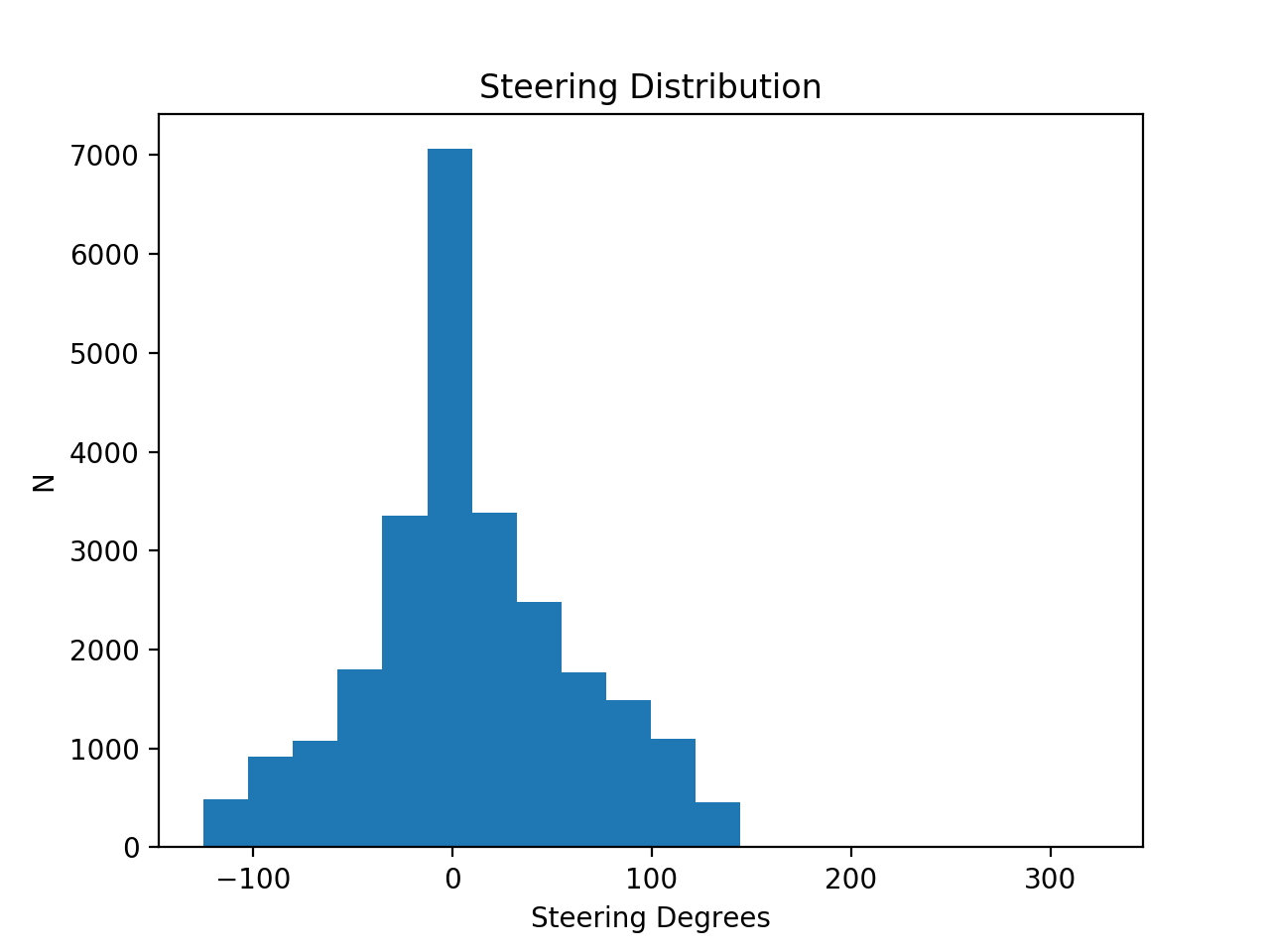}
    \caption{Steering Value Distribution}
    \label{fig:my_label}
\end{figure}
\begin{figure}[h!]
    \centering
    \includegraphics[scale=0.4]{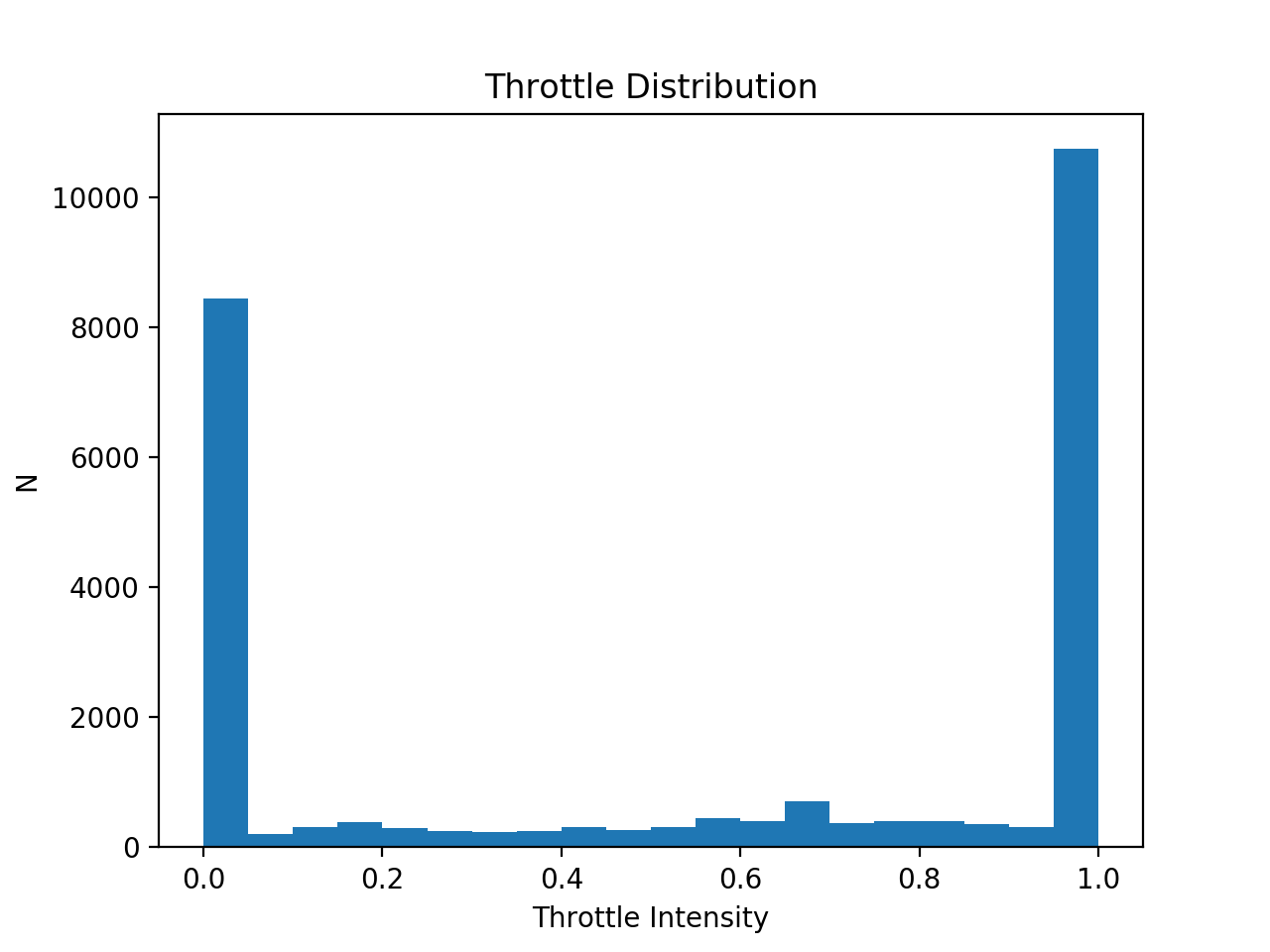}
    \caption{Throttle Value Distribution}
    \label{fig:my_label}
\end{figure}
\begin{figure}[h!]
    \centering
    \includegraphics[scale=0.4]{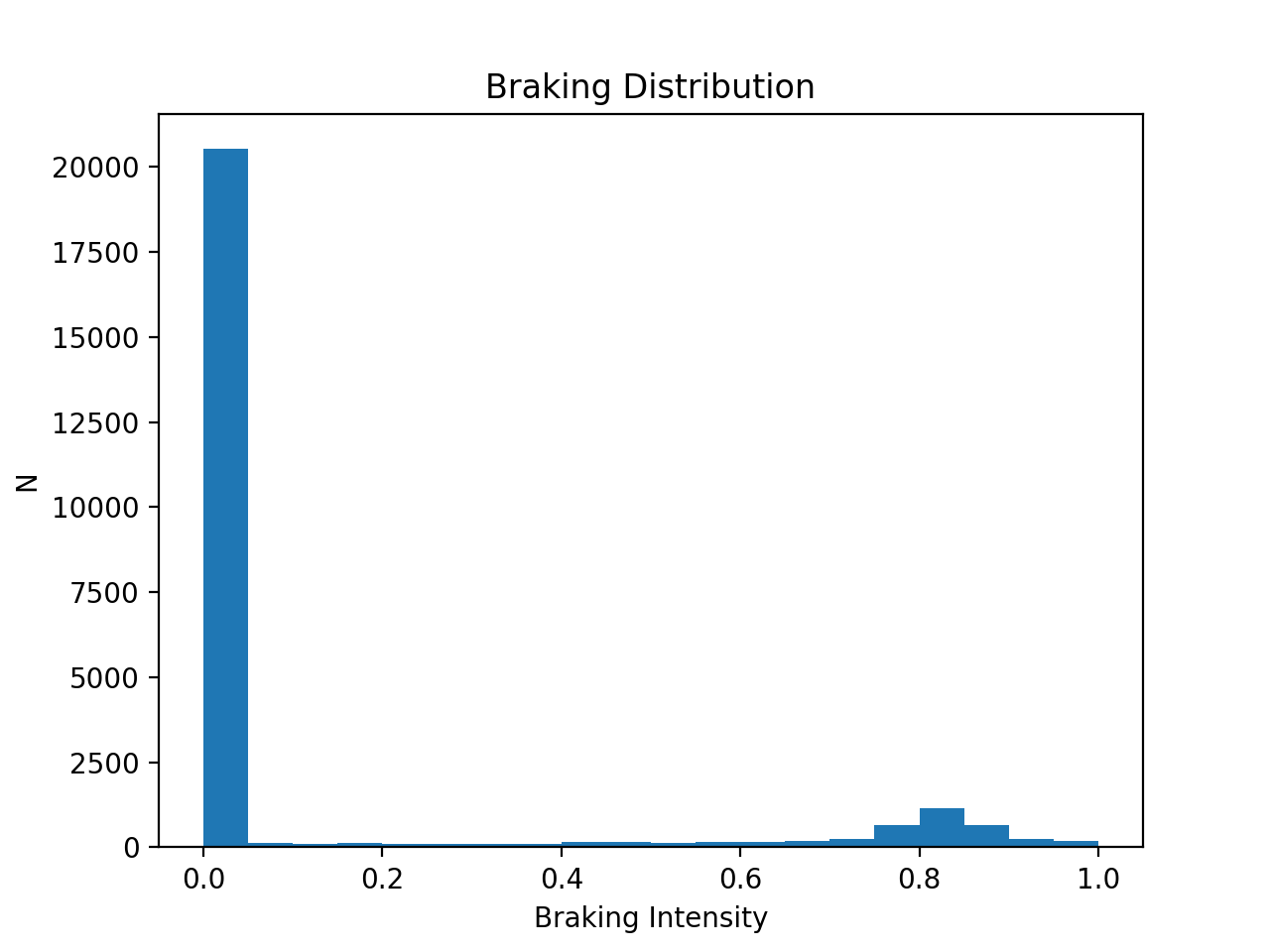}
    \caption{Braking Value Distribution}
    \label{fig:my_label}
\end{figure}
Steering, which the model has fit well, is almost perfectly normally distributed, and centered around zero. But braking and throttle are both extremely multimodal, with most of their mass on 0 or 1, which suggested that a regression model would have difficultly fitting these beyond fitting these distributions beyond noise. Altogether, this motivated a number of architecutral changes.

\subsection{Final Architecture: ResNet}
First, we aggressively downsampled our input images, from 3-channel RGB images at 1280x720 resolution, to 1-channel grayscale images at 320x180 resolution, to remove as much noise as possible while still retaining essential state information. Then, we applied a number of augmentation techniques at train time, including (1) randomly jittering each frame by 5 pixels (2) adding a random epsilon of Gaussian blur to each frame and (3) randomly flipping each frame along it's horizontal axis (and inverting it's corresponding steering value), all of which were aimed to improve generalization and prevent the model from fitting noise. Finally, we chose to reframe the problem as one of classification, between 36 values of steering (-180 degrees to 180 degrees, in 10 degree buckets) and 3 values for acceleration: full throttle, full braking, or neutral. We chose an 18-layer ResNet as our classification architecture, as we then believed that this task could be more difficult for a small convolutional network than we originally believed,
but still wanted to the flexibility of the network to degrade (via identity mappings) to a smaller network, if that were more ideal.
The final accuracy of our model was found to be \textbf{Final Acceleration Accuracy}: 0.86 \textbf{Final Steering Accuracy}: 0.58. For a result of the training loss of this model see our Figure 9 below. 

\begin{figure*}[h!]
    \centering
    \includegraphics[scale=0.4]{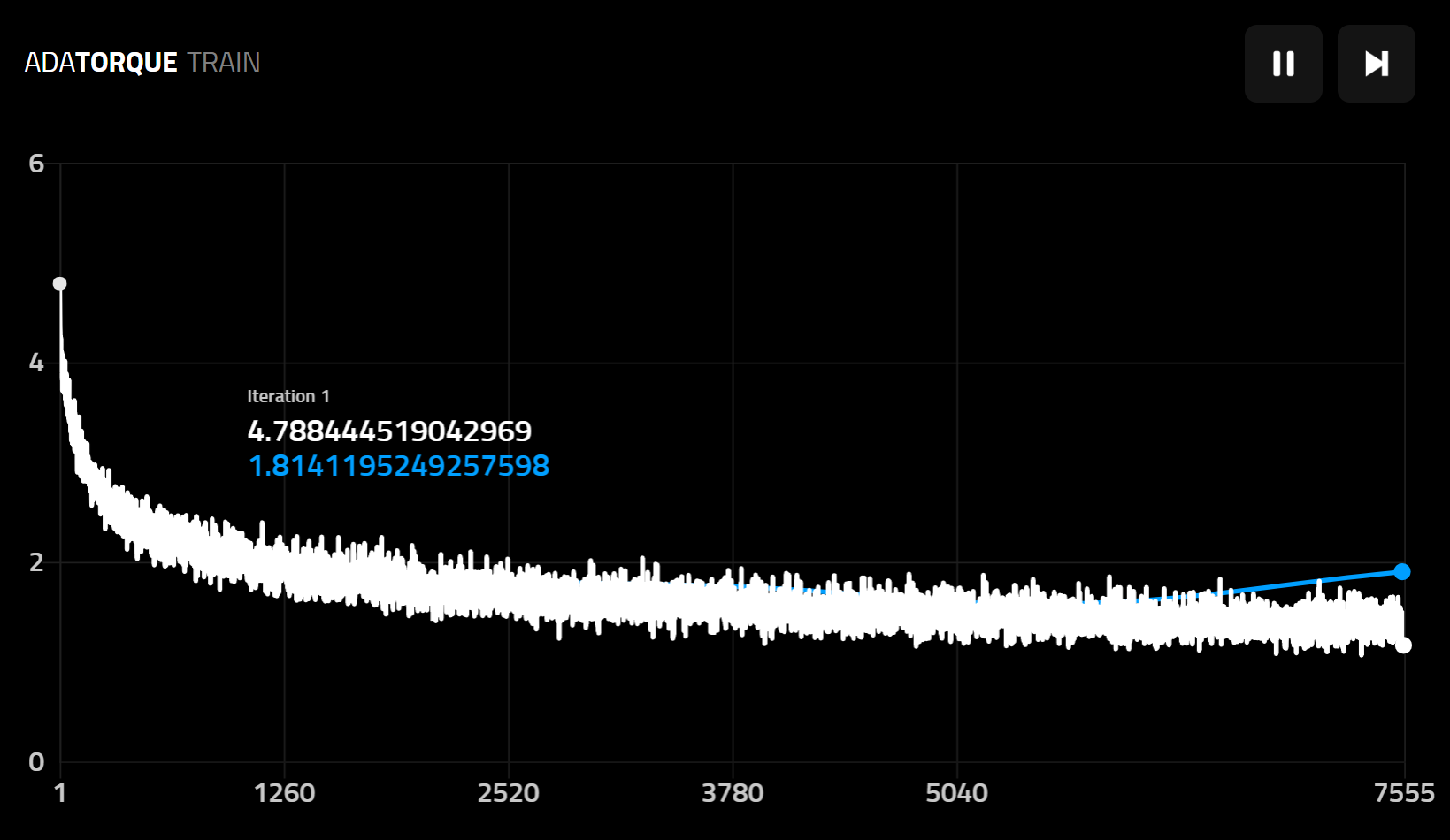}
    \caption{ResNet Model Training}
    \label{fig:my_label}
\end{figure*}

\pagebreak

\section{Conclusion} 

Our paper shows promising results for the ability of end-to-end CNN frameworks to learn complex handling policies for self-driving vehicles. Using a ResNet18 model, we were able to achieve a strong prediction for optimal steering, throttle and braking values.

As mentioned in our literature review, we may consider including LSTM-Conv layers into our model in order to create a more stateful driving policy. As Chi et al. point out in their paper, it is counter-intuitive to only focus on the current time frame in order to predict optimal handling policies. Rather, allowing the model to learn an intuition about past driving behavior may prove important for predicting future behavior. 

Along the same lines, our model inherently is biased towards predict policies which are fine-tuned for the particular vehicle that it was trained on. Ideally, we would train our neural network on a variety of different vehicles which exhibit distinct handling properties. By doing so, we may be able to create more general driving policies that can be applied to a variety of simulated vehicles. Perhaps we may also consider applying some of the research into recent Theory of Mind models within this context. That is to say, it seems plausible to create a meta-distribution over the different type of handling policies that apply to the unique sets of cars in the simulator (e.g extremely high-performing race cars, musclecars, SUVs...). By conditioning on the current type of car, one could then optimize a conditional self-driving policy that would implement more tailored handling rules. 

\footnotesize

\bibliographystyle{apalike}
\bibliography{sample}

\end{document}